\documentclass[review]{elsarticle}

\usepackage{lineno,hyperref}
\modulolinenumbers[5]

\journal{Theoretical Computer Science}

\usepackage{amssymb,amsthm}

\usepackage{url}

\usepackage{amsmath,amsfonts,stmaryrd,amssymb}
\usepackage{algorithm}

\usepackage{algorithmic}

\usepackage{booktabs}
\usepackage{pifont}

\usepackage{tikz}
\usetikzlibrary{arrows}
\usetikzlibrary{calc}

\newtheorem{theorem}{Theorem}[section]
\newtheorem{lemma}[theorem]{Lemma}

\newtheorem{definition}[theorem]{Definition}

\newtheorem{assumption}{Assumption}{\bfseries}{\itshape}

\renewcommand{\eqref}[1]{Eq.~(\ref{eq:#1})}

\newcommand{\tabref}[1]{Table \ref{tab:#1}}        
\newcommand{\secref}[1]{Section \ref{sec:#1}}
\newcommand{\thmref}[1]{Theorem \ref{thm:#1}}
\newcommand{\lemref}[1]{Lemma \ref{lem:#1}}

\newcommand{\asref}[1]{Assumption \ref{as:#1}}

\newcommand{\reals}{\mathbb{R}}

\newcommand{\half}{{\frac12}}
\newcommand{\ceil}[1]{{\lceil #1\rceil}}

\newcommand{\one}{\mathbb{I}}

\DeclareMathOperator*{\argmin}{argmin}
\DeclareMathOperator*{\argmax}{argmax}

\newcommand{\st}{\text{ s.t. }}

\newcommand{\cA}{\mathcal{A}}

\newcommand{\cF}{\mathcal{F}}
\newcommand{\cG}{\mathcal{G}}
\newcommand{\cH}{\mathcal{H}}

\newcommand{\cX}{\mathcal{X}}
\newcommand{\cY}{\mathcal{Y}}
\newcommand{\cZ}{\mathcal{Z}}

\usepackage{multirow}

\renewcommand{\citet}{\cite}
\renewcommand{\citep}{\cite}

\newcommand{\OPT}{\mathrm{OPT}}
\DeclareMathOperator{\minsec}{\phi}
\newcommand{\cost}{\mathfrak{cost}}
\newcommand{\vs}{\mathrm{VS}}

\newcommand{\pairs}{{\cX \times \cY}}
\newcommand{\nil}{\bot}
\newcommand{\val}{\mathrm{val}}
\newcommand{\cmax}{\cost_{\max}}

\newcommand{\crr}{r} 
\newcommand{\sscr}{r^{[2]}} 
\newcommand{\gsscr}{gr^{[2]}}

\newcommand{\tableresults}
{

\begin{tabular}{|l|l|l|l|l||l|l|l|}
\hline
\multicolumn{5}{|c|}{Test parameters} & \multicolumn{3}{|c|}{Results: $\cost(\cA)$}\\
\hline
Dataset & $f$ & \# communities, & $\sscr_\cost$ & $\gsscr_\cost$ & $u^f$ & $u_2^f$ & $u_3^f$ \\
&  & $|\cH|$ & &  &  &  & \\ 
\hline
\hline
Facebook & edge users &3, 100 & 5 & 5 & \textbf{52} & 255 & 157 \\\hline
Facebook &edge users &3, 100  & 100 & 100 & \textbf{148} & 5100 & 2722 \\\hline
Facebook &edge users &3, 100   &  1 & 100 & \textbf{49} & 52 & 2821 \\\hline

Facebook &edge users &10, 500& 5 & 5 & \textbf{231} & 256 & 242  \\\hline
Facebook &edge users &10, 500 & 100 & 100  & \textbf{4601} & 5101 & 4802  \\\hline
Facebook &edge users &10, 500 & 1 & 100 & \textbf{50} & 52 & 2915\\\hline

Facebook &version-space reduction & 3, 100 & 5 & 5 & \textbf{13} & 20 & 15 \\\hline
Facebook &version-space reduction & 3, 100  & 100 & 100 & \textbf{203} & 400 & 300 \\\hline
Facebook &version-space reduction & 3, 100 & 1 & 100 & \textbf{3} & 4 & 201 \\
\hline
Facebook &version-space reduction &10, 500  & 5 & 5 & \textbf{8} & 20 & 15   \\\hline
Facebook &version-space reduction &10, 500 & 100 & 100  & \textbf{105} & 400 & 300  \\\hline
Facebook &version-space reduction &10, 500 & 1 & 100 & \textbf{101} & 103 & 201\\
\hline
GR-QC & edge users &3, 100 & 5 & 5 & \textbf{51} &181 & 123\\\hline
GR-QC &edge users &3, 100  & 100 & 100 & \textbf{147} &3503 & 1833\\\hline
GR-QC &edge users &3, 100 & 1 & 100 & \textbf{51} & 53 & 2526\\
\hline
GR-QC &edge users &10, 500 & 5 & 5 & 246 & 260 & \textbf{245}\\\hline
GR-QC &edge users &10, 500&  100 & 100  & 4901 & 5200& \textbf{4900} \\\hline
GR-QC &edge users &10, 500& 1 & 100 & \textbf{49} & 52 & 3217 \\
\hline
GR-QC &version-space reduction & 3, 100 & 5 & 5 & \textbf{10}& 20 & 15 \\\hline
 GR-QC &version-space reduction & 3, 100 & 100 & 100 & \textbf{106} & 400 & 300  \\\hline
GR-QC &version-space reduction & 3, 100 & 1 & 100 & \textbf{3} & 400 & 300  \\
\hline

GR-QC &version-space reduction &10, 500 & 5 & 5 & \textbf{15}& 16& \textbf{15} \\\hline
GR-QC &version-space reduction &10, 500  & 100 & 100 & \textbf{300} & 301 & \textbf{300}  \\\hline
GR-QC &version-space reduction &10, 500  & 1 & 100 & \textbf{3} &201 & 300 \\
\hline
\end{tabular}

}

\newcommand{\tablecostslearn}
{
\begin{table}[h]
\center
\begin{tabular}{l|l|l}
 & $h_j$& $h_i$, $i \neq j$\\
\hline
$a_j$ & 1  & 2\\
\hline
$b^t_j$  & 3& $l_{i,j}^t+1$
\end{tabular}
\caption{The values of functions in $\cH$ for the proof of \thmref{lowerboundlearning}}
\label{tab:costslearn}
\end{table}
}

\newcommand{\tablecostsgen}
{
\begin{table}[h]
\center
\begin{tabular}{l|l|l|l}
& & $y=0$& $y= 1$\\
\hline
$a_i$ & $\cost(a_i,y)$ & $c_1$&$c_1$ \\
 & $f((a_i,y))$ & $0$&$Q$\\
\hline
$b_i$ & $\cost(b_i,y)$ & $c_1$& $c_2$\\
& $f((b_i,y))$ & $Q$ & $Q$.\\
\hline
$c$ & $\cost(c,y)$ & $c_3$ & $c_4$\\
& $f((c,y))$ & 0 & 0\\
\end{tabular}
\caption{The cost function and the objective function in the proof of \thmref{lowerboundgeneral}}
\label{tab:costsgen}
\end{table}
}

\begin{document}
\begin{frontmatter}

\title{Submodular Learning and Covering\\ with Response-Dependent Costs}

\author{Sivan Sabato\corref{mycorrespondingauthor}}
\address{Ben-Gurion University of the Negev, Beer Sheva 8499000, Israel.}
\cortext[mycorrespondingauthor]{Corresponding author}
\ead{sabatos@cs.bgu.ac.il}

\begin{abstract}
We consider interactive learning and covering problems, in a setting where actions may incur different costs, depending on the response to the action. 
We propose a natural greedy algorithm for response-dependent costs. We bound the approximation factor of this greedy algorithm in active learning settings as well as in the general setting. We show that a different property of the cost function controls the approximation factor in each of these scenarios.
We further show that in both settings, the approximation factor of this greedy algorithm is near-optimal among all greedy algorithms. Experiments demonstrate the advantages of the proposed algorithm in the response-dependent cost setting.
\end{abstract}

\begin{keyword}
Interactive learning, submodular functions, outcome costs
\end{keyword}

\end{frontmatter}


\section{Introduction}

We consider interactive learning and covering problems, a term introduced in \citet{GuilloryBi10}. In these problems, there is an algorithm that interactively selects actions and receives a response for each action. Its goal is to achieve an objective, whose value depends on the actions it selected, their responses, and the state of the world. The state of the world, which is unknown to the algorithm, determines the response to each action. The algorithm incurs a cost for every action it performs. The goal is to have the total cost incurred by the algorithm as low as possible.

Many real-world problems can be formulated as interactive learning and covering problems. 
For instance, in pool-based active learning problems \citep{MccallumNi98,Dasgupta04},
each possible action is a query of the label of an example, and the goal is to identify the correct mapping from examples to labels out of a given set of possible mappings. Another example is maximizing the influence of marketing in a social network \citep{GuilloryBi10}. In this problem, an action is a promotion sent to specific user, and the goal is to make sure all users of a certain community are affected by the promotion, either directly or via their friends. There are many other applications for interactive algorithms. As additional examples, consider interactive sensor placement \cite{GolovinKr11} and document summarization \cite{LinBi11} with interactive user feedback.

Interactive learning and covering problems cannot be solved efficiently in general \citep{NemhauserWoFi78,Wolsey82}. Nevertheless, many such problems can be solved near-optimally by efficient algorithms, when the functions that map the sets of actions to the total reward are \emph{submodular}.

It has been shown in several settings, that a simple greedy algorithm pays a near-optimal cost when the objective function is submodular (e.g.,~\cite{GuilloryBi10,GolovinKr11,CuongLeYe14}). Many problems naturally lend themselves to a submodular formulation. 
For instance, a pure covering objective is usually submodular, and so is an objective in which diversity is a priority, such as finding representative items in a massive data set \cite{MirzasoleimanKaSaKr13}.  
Active learning can also be formalized as a submodular interactive covering objective, leading to efficient algorithms \cite{Dasgupta04,GolovinKr11,GuilloryBi10,GonenSaSh13b}.

Interactive learning and covering problems have so far been studied mainly under the assumption that the cost of the action is known to the algorithm before the action is taken. In this work we study the setting in which the costs of actions depend on the outcome of the action, which is only revealed by the observed response. This is the case in many real-world scenarios. For instance, consider an active learning problem, where the goal is to learn a classifier that predicts which patients should be administered a specific drug. Each action in the process of learning involves administering the drug to a patient and observing the effect. In this case, the cost (poorer patient health) is higher if the patient suffers adverse effects. Similarly, when marketing in a social network, an action involves sending an ad to a user. If the user does not like the ad, this incurs a higher cost (user dissatisfaction) than if they like the ad. 

We study the achievable approximation guarantees in the setting of
response-dependence costs, and characterize the dependence of this
approximation factor on the properties of the cost function.  We propose a
natural generalization of the greedy algorithm of \citet{GuilloryBi10} to the
response-dependent setting, and provide two approximation guarantees. The
first guarantee holds whenever the algorithm's objective describes an active
learning problem. We term such objectives \emph{learning objectives}. The
second guarantee holds for general objectives, under a mild condition. In each
case, the approximation guarantees depend on a property of the cost function,
and we show that this dependence is necessary for any greedy algorithm. Thus,
this fully characterizes the relationship between the cost function
and the approximation guarantee achievable by a greedy algorithm.
We further report experiments that demonstrate the achieved cost improvement.

Response-dependent costs has been previously studied in specific cases of
active learning, assuming there are only two possible labels
\citep{SabatoSaSr13,SabatoSaSr15,SaettlerLaCi14,SaettlerLaCi15}.  In
\citet{KapoorHoBa07} this setting is also mentioned in the context of active
learning. Our work is more general: First, it addresses general objective
functions and not only specific active learning settings. Our results indicate that the active learning setting and the general setting are inherently different. Second, our analysis is not limited to
settings with two possible responses. 
As we show below, a straightforward generalization of previous guarantees for two responses to more than two responses results in loose bounds. 
 We thus develop new proof techniques that allow deriving tighter bounds.

The paper is structured as follows. Definitions and preliminaries are given in \secref{setting}. We show a natural generalization of the greedy algorithm to response-dependent costs in \secref{gen}. We provide tight approximation bounds for the greedy algorithm, and matching lower bounds, in \secref{tight}. Experiments are reported in \secref{exp}. We conclude in \secref{conclusion}.

\section{Definitions and Preliminaries}\label{sec:setting}
For an integer $n$, denote $[n] := \{1,\ldots,n\}$. A set function $f:2^\cZ \rightarrow \reals$ is \emph{monotone} (non-decreasing) if 
\[
\forall A\subseteq B \subseteq \cZ,\quad f(A) \leq f(B). 
\]
Let $\cZ$ be a domain, and let $f:2^\cZ \rightarrow \reals_+$ be a set function. Define, for any $z\in\cZ,A\subseteq \cZ$, 
\[
\delta_f(z\mid A) := f(A\cup \{z\}) - f(A).
\]
 $f$ is \emph{submodular} if 
\[
\forall z \in \cZ, A\subseteq B \subseteq \cZ,\quad \delta_f(z \mid A) \geq \delta_f(z \mid B).
\]

Assume a finite domain of actions $\cX$ and a finite domain of responses $\cY$.
For simplicity of presentation, we assume that there is a one-to-one mapping between world states and mappings from actions to responses. Thus the states of the world are represented by the class of possible mappings $\cH \subseteq \cY^\cX$.
Let $h^* \in \cH$ be the true, unknown, mapping from actions to responses.
Let $S \subseteq \pairs$ be a set of action-response pairs.

We consider algorithms that iteratively select a action $x \in \cX$ and get the response $h^*(x)$, where $h^* \in \cH$ is the true state of the world, which is unknown to the algorithm. For an algorithm $\cA$, let $S^h[\cA]$ be the set of pairs collected by $\cA$ until termination if $h^* = h$. Let $S^h_t[\cA]$ be the set of pairs collected by $\cA$ in the first $t$ iterations if $h^* = h$. In each iteration, $\cA$ decides on the next action to select based on responses to previous actions, or it decides to terminate. $\cA(S) \in \cX \cup \{\nil\}$ denotes the action that $\cA$ selects after observing the set of pairs $S$, where $\cA(S) = \nil$ if $\cA$ terminates after observing $S$. 

Each time the algorithm selects an action and receives a response, it incurs a cost, captured by a cost function 
$\cost:\cX \times \cY \rightarrow \reals_+$. If $x \in \cX$ is selected and the response $y \in \cY$ is received, the algorithm pays $\cost(x,y)$. Denote 
\[
\cost(S) = \sum_{(x,y) \in S} \cost(x,y).
\]
The total cost of a run of the algorithm, if the state of the world is $h^*$, is thus $\cost(S^{h^*}[\cA])$. For a given $\cH$, define the \emph{worst-case cost} of $\cA$ by 
\[
\cost(\cA) := \max_{h \in \cH} \cost(S^h[\cA]).
\] Let $Q > 0$ be a threshold, and let $f:2^\pairs \rightarrow \reals_+$ be a monotone non-decreasing submodular objective function. The goal of the interactive algorithm is to collect pairs $S$ such that $f(S) \geq Q$, while minimizing $\cost(\cA)$.

Guillory and Bilmes \cite{GuilloryBi10} consider a setting 
 in which instead of a single global $f$, there is a set of monotone non-decreasing objective functions
\[
\cF_{\cH} = \{f_h:2^\pairs \rightarrow \reals_+ \mid h \in \cH\},
\] and the value $f_h(S)$, for $S \subseteq \pairs$, represents the reward obtained by the algorithm if $h^* = h$. They define a surrogate set function $\bar{F}:2^\pairs \rightarrow \reals_+$ as follows: 
\begin{equation}\label{eq:fbar}
\bar{F}(S) := \frac{1}{|\cH|}\Big(Q|\cH \setminus \vs(S)| + \sum_{h \in \vs(S)}\min(Q,f_h(S))\Big).
\end{equation}
Here $\vs(S)$ is the \emph{version space} induced by $S$ on $\cH$, defined by 
\[
\vs(S) = \{ h \in \cH \mid \forall (x,y) \in S, y = h(x)\}. 
\]
They show that if the algorithm obtains $\bar{F}(S) \geq Q$, then it has also
obtained $f_{h^*}(S) \geq Q$. In the other direction, if the algorithm obtains
$f_{h^*}(S) \geq Q$ and \emph{knows} that it has done so (equivalently, the algorithm obtains $f_{h}(S) \geq Q$ for all $h \in \vs(S)$), then it has also
obtained $\bar{F}(S) \geq Q$. In other words, obtaining $\bar{F}(S) \geq Q$ is equivalent to a guarantee of the algorithm that $f_{h^*}(S) \geq Q$. 

It is shown in \cite{GuilloryBi10} that if all the functions in $\cF_\cH$ are monotone and submodular then so is $\bar{F}$. Thus our setting of a single objective function can be applied to the setting of \cite{GuilloryBi10} as well.

Let $\alpha \geq 1$. An interactive algorithm $\cA$ is an \emph{$\alpha$-approximate} \emph{greedy algorithm} for \emph{utility function} \mbox{$u:\cX \times 2^\pairs \rightarrow \reals_+$}, if the following holds: For all $S \subseteq \pairs$, if $f(S) \geq Q$ then $\cA(S) = \nil$, and otherwise, $\cA(S) \in \cX$ and 
\[
u(\cA(S),S) \geq \frac{1}{\alpha}\max_{x \in \cX} u(x,S).
\]
Competitive guarantees are generally better for $\alpha$-approximate-greedy algorithms with $\alpha$ closer to $1$ \cite{GolovinKr11}. However, because of computational issues or other practical considerations, it is not always feasible to implement a $1$-greedy algorithm. Thus, for full generality, we analyze  also $\alpha$-greedy algorithms for $\alpha > 1$. 

Let $\OPT := \min_{\cA} \cost(\cA)$, where the minimum is taken over all interactive $\cA$ that obtain $f(S) \geq Q$ at termination, for all possible $h^* \in \cH$. If no such $\cA$ exist, define $\OPT = \infty$.

In \cite{GuilloryBi10} it is assumed that costs are not response-dependent, 
thus $\cost(x,y) \equiv \cost(x)$, and a greedy algorithm is proposed, based on the following utility function:
\begin{equation}\label{eq:utilitygbone}
u(x,S) :=\min_{h \in \vs(S)}\frac{\delta_{\bar{F}}((x,h(x)) \mid S)}{\cost(x)}.
\end{equation}
It is shown that for functions $f$ with an integer range, and for an integer $Q$, this algorithm obtains $f(S) \geq Q$ with a worst-case cost of at most \mbox{$\textrm{GCC}(\ln(Q|\cH|)+1)$}, 
where $\textrm{GCC}$ is a lower bound on $\OPT$.
In \citet{GolovinKr11}, a different greedy algorithm and analysis guarantees a worst-case cost of $\alpha (\ln(Q) +1 )\cdot \OPT$ for adaptive submodular objectives and $\alpha$-approximate greedy algorithms. The factor of $\ln(Q)$ cannot be substantially improved by an efficient algorithm, even for non-interactive problems \citep{Wolsey82,Feige98}.

\subsection{An example}\label{sec:example}

We give a concrete example of a problem that can be formalized using the definitions above. Consider for instance a problem of promotion in a social network, where users form a graph based on friendships. Each user belongs to one community, and the goal is to contact $Q$ users who have at least one friend in a different community than their own. Each action in $\cX$ is mapped to a single network user, and refers to contacting the user by sending a promotional email. The response of the user identifies the user's community. The possible states of the world $\cH$ correspond to possible mappings of the users into communities. 

To define the objective function $f$, we first define a set of hypothesis-dependent objective functions 
$\cF_{\cH} := \{f_h:2^\pairs \rightarrow \reals_+ \mid h \in \cH\}$, where $f_h(S)$ is the number of users with friends in a different community that have been contacted in $S$. Formally,
\begin{align*}
f_h(S) = | \{x \in \cX \,\mid\, &(x,h(x)) \in S \text{ and }\\
& (\exists z \in \cX, h(z) \neq h(x) \text{ and users } x \text{ and } z \text{ are friends}) \}|.
\end{align*}
Clearly, $f_h(S)$ is monotone and submodular. The global function $f$ is set to be equal to $\bar F$, as defined in \eqref{fbar}.  A greedy algorithm that uses the utility function $u$ selects, at each round, the user that increases $f$ the most relative to the cost of contacting that user. 

In the setting studied in previous works, where $\cost(x,y) \equiv \cost(x)$, the cost of contacting a user depends only on the user but not on the community it belongs to. This does not take into account possible differences between users, which can only be identified after contacting them. For instance, if one of the communities is a community of users who do not like promotional emails, they might mark the sender as ``Spam'', thus imposing a high cost on the sender. An algorithm that ignores this might incur higher costs, since it does not attempt to avoid such users. In the next section we propose a utility function that takes the dependence of the cost on the responses into account. 

Our results below show that the term controlling the ability to well-approximate the optimal solution under response-dependent costs is the ratio between the largest cost, $\max_{y\in \cY} \cost(x,y)$, and the second-smallest cost in the multiset $\{\cost(x,y) \mid y \in \cY\}$. For instance, consider the following scenario: Suppose there is a single possible response that is cheap (e.g., a user redeems the promotion), while other responses are all similarly expensive (e.g., a user is unhappy about the promotion and expresses it in one of several different ways, all of which decrease the seller's reputation). In this case, this ratio is $1$, implying that no additional deterioration of the approximation factor is incurred by the fact that the costs are response-dependent.  As we show below, straightforward generalizations of previous work to response-dependent costs would give in this case an approximation factor that grows with the cost of the expensive action.

\section{Generalizing to response-dependent costs}\label{sec:gen}

The results of \cite{GuilloryBi10} can be generalized to the response-dependent cost setting using the \emph{cost ratio} of the problem, denoted $\crr_\cost$ and defined by:
\[
\crr_\cost:= \max_{x \in \cX} \frac{\max_{y \in \cY} \cost(x,y)}{\min_{y \in \cY} \cost(x,y)}.
\]
Consider a generalized version of the utility function $u$ given in \eqref{utilitygbone}:
\begin{equation}\label{eq:utilitygb}
u(x,S) :=\min_{h \in \vs(S)}\frac{\delta_{\bar{F}}((x,h(x)) \mid S)}{ \cost(x,h(x))}.
\end{equation}
Setting $\overline{\cost}(x) := \min_{y \in \cY} \cost(x,y)$, we have $\cost \leq \crr_\cost \cdot \overline{\cost}$. Using this fact, an approximation guarantee of $\crr_{\cost}\cdot \OPT(\ln(Q|\cH|)+1)$ is immediate for a greedy algorithm which uses the utility function in \eqref{utilitygb} with a response-dependent cost. Similarly, it is immediate to derive an approximation factor of $\crr_\cost\cdot \alpha (\ln(Q) + 1) \OPT$ in the setting of \cite{GolovinKr11}. 
However, in this work we show that this natural derivation is loose: We provide  tighter approximation bounds, which can be finite even if $\crr_\cost$ is infinite. 
Our results below hold for any function $f$ which satisfies the following standard assumptions (e.g.~\cite{GolovinKrRa10}).
\begin{assumption}\label{as:fterms}
Let $f:2^\pairs \rightarrow \reals_+$, $Q >0$, $\eta > 0$. Assume that $f$ is submodular and monotone, $f(\emptyset) = 0$, and that for any $S \subseteq \pairs$, if $f(S) \geq Q -\eta$ then $f(S) \geq Q$.
\end{assumption}

The assumption regarding $\eta$ is a standard generalization of the more restrictive assumption that $f$ returns integer values. Such an $f$ has $\eta = 1$, but it is also possible to have $\eta = 1$ for functions that return some fractional values. Our guarantees depend only on the ratio $Q/\eta$, hence are invariant to a linear scaling of $f$. 

We analyze a greedy algorithm that selects an element maximizing (or approximately maximizing) the following utility function:
\begin{equation}\label{eq:uf}
u^f(x,S) := \min_{h \in \vs(S)}\frac{\delta_{\min(f,Q)}((x,h(x)) \mid S)}{\cost(x,h(x))}.
\end{equation}
Note that $u^{\bar{F}}$ is equal to the function $u$ defined in \eqref{utilitygb}.

\section{Tight approximation bounds for the generalized greedy algorithm}\label{sec:tight}
We provide approximation guarantees for the greedy algorithm which maximizes the utility function in \eqref{uf}, under two types of objective functions. The first type captures active learning settings, while the second type is more general. Our results show that objective functions for active learning have better approximation guarantees than general objective functions. 

In \secref{learningobj} we show an approximation guarantee for objectives that are useful for active learning, which we term \emph{learning objectives}. We give a matching lower bound in \secref{learninglower}.
In \secref{generalobj} we consider general monotone submodular objective functions. We give a matching lower bound in \secref{generallower}.

Our guarantees hold for objective functions $f$ that satisfy the following property, which we term \emph{consistency-aware}. This property requires that the function gives at least $Q$ to any set of action-response pairs that are inconsistent with $\cH$. 
\begin{definition}[consistency-aware]\label{def:inconsistentweak}
A function $f:2^\pairs \rightarrow \reals_+$ is \emph{consistency-aware} for threshold $Q > 0$ if for all $S \subseteq \pairs$ such that $\vs(S) = \emptyset$, $f(S) \geq Q$.
\end{definition}
Note that the definition is concerned with the value of $f$ only on inconsistent sets $S$, which the algorithm never encounters. Therefore, it suffices that there exist an extension of $f$ to these sets that is consistent with all the other requirements from $f$. The function $\bar{F}$ defined in \eqref{fbar} is consistency-aware. In addition, a construction similar to $\bar{F}$, with non-uniform weights for the possible mappings, is also consistency-aware. Such
a construction is sometimes more efficient to compute than the uniform-weight construction. For instance, as shown in  \citep{GonenSaSh13b}, non-uniform weights allow a more efficient computation when the mappings
represent linear classifiers with a margin. In general, any objective $f$ can be made consistency aware using a simple transformation such as $\bar{F}$. Thus our results are applicable to a diverse class of problems.

\subsection{An approximation upper bound for learning objectives}\label{sec:learningobj}
Active learning is an important special case of interactive learning. In active learning, the only goal is to discover information on the identity of $h^*$. We term functions that represent such a goal \emph{learning objectives}. 
\begin{definition}\label{def:learning}
A function $f:2^\pairs \rightarrow \reals_+$ is a \emph{learning objective}  for $\cH$ if $f(S) = g(\vs(S))$ where $g$ is a monotone non-increasing function.
\end{definition}
It is easy to see that all learning objectives $S \mapsto f(S)$ are monotone non-decreasing in $S$. In many useful cases, they are also submodular. In noise-free active learning, where the objective is to exactly identify the correct mapping $h^*$, one can use the learning objective 
\[
f(S) := 1 - |\vs(S)|/|\cH|, 
\]
with $Q = 1-1/|\cH|$.
This is the \emph{version-space reduction} objective function \citep{GolovinKr11,GuilloryBi10}. 

In
\citet{GolovinKrRa10}, the problems of noise-aware active learning, and its generalization to
Equivalence Class Determination, are considered. In this generalization, there is some partition of $\cH$, and the goal is to identify
the class to which $h^*$ belongs. The objective function proposed by
\citet{GolovinKrRa10}, measures the weight of pairs in $\vs(S)$ which include two mappings that belong to different classes. This function is also a learning
objective. In \citet{CuongLeYe14} the \emph{total generalized version-space reduction} function is proposed. This function is also a learning objective. More generally, consider a set of structures $\cG \subseteq 2^\cH$, where the goal is to disqualify these structures from the version space, by proving  that at least one of the mappings in this structure cannot be the true $h^*$. In this case one can define the submodular learning objective 
\[
f(S) := w(\cG) - w(\cG \cap 2^{\vs(S)}), 
\]
where $w$ is a modular weight function on $\cG$, and $Q = w(\cG)$. For instance, if $\cG$ is the set of pairs from different equivalence classes in $\cH$, this is the Equivalence Class Determination objective. If $\cG$ is a set of triplets from different equivalence classes, this encodes an objective of reducing the uncertainty on the identity of $h^*$ to at most two equivalence classes.

We show that for learning objectives, the approximation factor for a greedy algorithm that uses $u^f$ depends on a new property of the cost function, which we term the \emph{second-smallest cost ratio}, denoted by $\sscr_\cost$. 
For $x \in \cX$, let $\minsec(x)$ be the second-smallest value in the multiset \mbox{$\{ \cost(x,y) \mid y \in \cY\}$}. Define
\[
\sscr_\cost := \max_{x\in \cX,y \in \cY}\frac{\cost(x,y)}{\minsec(x)}.
\]

\begin{theorem}\label{thm:worstcase}
Let $f:2^\pairs \rightarrow \reals_+,Q >0,\eta > 0$ such that \asref{fterms} holds.
Let $\cA$ be an $\alpha$-approximate greedy algorithm for the utility function $u^f$. If $f$ is a learning objective, then 
\[
\cost(\cA) \leq \sscr_\cost \cdot \alpha   (\ln(Q/\eta)+1)\OPT.
\]
\end{theorem}

The ratio between the trivial bound that depends on the cost ratio
$\crr_\cost$, mentioned in \secref{setting}, and this new bound, is
$\crr_\cost/\sscr_\cost$, which is unbounded in the general case: for
instance, if each action has one response which costs $1$, and the other
responses cost $M \gg 1$, then $\crr_\cost = M$ but $\sscr_\cost = 1$.  Whenever $|\cY| = 2$, $\sscr_\cost =1$.
Thus, the approximation factor of the greedy algorithm for any binary active
learning problem is independent of the cost function. This coincides with the
results of \cite{SaettlerLaCi14,SaettlerLaCi15} for active learning with
binary labels. If $|\cY| > 2$, then the bound is smallest when
$\sscr_\cost = 1$, which would be the case if for each action there is one
preferred response which has a low cost, while all other responses have the
same high cost. For instance, this could be the case in a marketing application, in which the
action is to recommend a product to a user, and the response is either buying
the product (a preferred response), or not buying it, in which case additional
feedback could be provided by the user, but the cost (user dissatisfaction)
remains the same regardlesss of that feedback.

To prove \thmref{worstcase}, we use the following property of learning objectives: For such objectives, there exists an optimal algorithm (that is, one that obtains $\OPT$) that only selects actions for which at least two responses are possible given the action-response pairs observed so far. Formally, we define \emph{bifurcating} algorithms.
Denote the set of possible responses for $x$ given the history $S$ by 
\[
\cY_\cH(x,S) := \{ h(x) \mid h \in \vs(S)\}. 
\]
We omit the subscript $\cH$ when clear from context.
\begin{definition}
An interactive algorithm $\cA$ is \emph{bifurcating} for $\cH$ if for all $t$ and $h \in \cH$, 
\[
|\cY_\cH(\cA(S^h_{t}[\cA]),S^h_{t}[\cA])| \geq 2.
\]
\end{definition}

\begin{lemma}\label{lem:bifurcating}
For any learning objective $f$ for $\cH$ with an optimal algorithm, there exists an optimal algorithm for $f,\cH$ which is bifurcating.
\end{lemma}

\begin{proof}
Let $\cA$ be an optimal algorithm for $f$. Suppose there exists some $t,h$ such that 
$\cY(x_0,S^h_{t-1}[\cA]) = \{y_0\}$
 for some $y_0 \in \cY$, where $x_0 := \cA(S^h_{t-1}[\cA])$. Let $\cA'$ be an algorithm that selects the same actions as $\cA$, except that it skips the action $x_0$ it if has collected the pairs $S_{t-1}^h[\cA]$. That is, $\cA'(S) = \cA(S)$ for $S \nsupseteq S^h_{t-1}[\cA]$, and 
\[
\cA'(S) = \cA(S\cup \{(x_0,y_0)\}) 
\]
for $S \supseteq S^h_{t-1}$.
Since
\[
\vs(S) = \vs(S\cup \{(x_0,y_0)\}),
\]
and $\cA$ is a learning objective, $\cA'$ obtains $Q$ as well, at the same cost of $\cA$ or less. By repeating this process a finite number of steps, we can obtain an optimal algorithm for $\cH$ which is bifurcating.
\end{proof}

The following lemma is the crucial step in proving \thmref{worstcase}, and will also be used in the proof for the more general case below. 
The lemma applies to general consistency-aware functions. It can be used for learning objectives, because all learning objectives with a finite $\OPT$ are consistency-aware: Suppose that $f$ is a learning objective, and let $S \subseteq \pairs$ such that $\vs(S) = \emptyset$. For any $h \in \cH$, denote 
\[
S^h_* := \{(x,h(x)) \mid x \in \cX\}.
\]
 We have $\vs(S^h_*) \supseteq \vs(S)$, therefore, since $f$ is a learning objective, $f(S) \geq f(S^h_*)$. Since $\OPT$ is finite, $f(S^h_*) \geq Q$. Therefore $f(S) \geq Q$. Thus $f$ is consistency-aware.

\begin{lemma}\label{lem:onestep}
Let $f,Q,\eta$ which satisfy \asref{fterms} such that $f$ is consistency-aware.
Let $\cA$ be an interactive algorithm that obtains $f(S) \geq Q$ at termination. Let $\gamma = \sscr_\cost$ if $\cA$ is bifurcating, and let $\gamma = \crr_\cost$ otherwise.
then
\[
\exists x\in \cX \st u^f(x,\emptyset) \geq \frac{Q}{\gamma\cdot \cost(\cA)}. 
\]
\end{lemma}

\begin{proof}
Denote for brevity $\delta \equiv \delta_{\min(f,Q)}$.
Define $\bar{\cH} := \cY^\cX$.
Consider an algorithm $\bar{\cA}$ such that for any $S$ that is consistent with some $h\in \cH$ (that is $\vs(S) \neq \emptyset$), $\bar{\cA}(S) = \cA(S)$, and $\bar{\cA}(S) = \nil$ otherwise. 
Since $f$ is consistency-aware, we have $f(S^h[\bar\cA]) \geq Q$ for all $h \in \bar{\cH}$.

Consider a run of $\bar \cA$, and denote the pair in iteration $t$ of this run by $(x_t,y_t)$. Denote $S_{t} = \{(x_i,y_i) \mid i \leq t\}$. Choose the run such that in each iteration $t$, the response $y_t$ is in $\argmin_{y \in \cY}\delta(x_t,y\mid S_{t-1})/\cost(x_t,y)$. Let $T$ be the length of the run until termination.

Denote $\psi := \max_{ h \in \bar{\cH}} \cost(S^h[\bar \cA])$, the worst-case cost of $\bar \cA$ over $\bar \cH$.
We have
\begin{align*}
Q/\psi &\leq f(S_T)/\cost(S_T) \\
&= \frac{\sum_{t\in[T]} (f(S_t) - f(S_{t-1}))}{\sum_{t\in [T]} \cost(x_t,y_t)}\\
&= \frac{\sum_{t\in [T]} \delta((x_t,y_t)\mid S_{t-1})}{\sum_{t \in [T]} \cost(x_t,y_t)} \\
&\leq \max _{t\in [T]} \left(\delta((x_t,y_t)\mid S_{t-1})/\cost(x_t,y_t)\right),
\end{align*}
where we used $f(\emptyset) = 0$ in the second line.
Thus there exists some $t\in [T]$ such that 
$
Q/\psi \leq \delta((x_t,y_t)\mid S_{t-1})/\cost(x_t,y_t).
$

Therefore
\begin{align}
u(x_t,\emptyset) &= \min_{y \in \cY} \delta((x_t,y)\mid \emptyset)/\cost(x_t,y) \notag \\
&\geq \min_{y \in \cY} \delta((x_t,y)\mid S_{t-1})/\cost(x_t,y) \notag\\
&= \delta((x_t,y_t)\mid S_{t-1})/\cost(x_t,y_t) \geq Q/\psi.\label{eq:costbar}
\end{align}
The second line follows from the submodularity of $f$. The third line follows from the definition of $y_t$.

To prove the claim, we have left to show that \mbox{$\psi \leq \sscr \cdot \cost(\cA)$}.  
Consider again a run of $\bar \cA$. If all observed pairs are consistent with some $h \in \cH$, then $\bar \cA$ and $\cA$ behave the same. Hence $\cost(S^h[\bar \cA]) = \cost(S^h[\cA])$. 
Now, consider $h \in \bar{\cH} \setminus \cH$. By the definition of $\bar \cA$, $S^h[\bar \cA]$ is a prefix of $S^h[\cA]$. Let $T = |S^h[\bar \cA]|$ be the number of iterations until $\bar \cA$ terminates. Then $S_{T-1}^h[\bar \cA]$ is consistent with some $h' \in \cH$.

Let $x_T$ be the action that $\cA$ and $\bar \cA$ select at iteration $T$, and let $h' \in \cH$ which is consistent with $S_{T-1}^h[\bar \cA]$, and incurs the maximal possible cost in iteration $T$. Formally, $h'$ satisfies 
\[
h'(x_T) \in \argmax_{y \in \cY_\cH(x_T,S_{T-1}^h[\cA])}  \cost(x_T,y).
\]
 
Now, compare the run of $\bar \cA$ on $h$ to the run of $\cA$ on $h'$. 
In the first $T-1$ iterations, the algorithms observe the same pairs.
In iteration $T$, they both select $x_T$. $\bar \cA$ observes $h(x_T)$, while $\cA$ observes $h'(x_T)$. $\bar \cA$ terminates after iteration $T$. Hence
\begin{align*}
\cost(S^h[\bar \cA]) &= \cost(S_{T-1}^h[\cA]) + \cost(x_T,h(x_T))\\
&= \cost(S_T^{h'}[\cA]) - \cost(x_T,h'(x_T)) + \cost(x_T,h(x_T)).
\end{align*}
Consider two cases: (a) $\cA$ is not bifurcating. Then $\gamma = \crr$, and so 
\[
\cost(x_T,h(x_T)) \leq \gamma \cost(x_T,h'(x_T)).
\]
 (b) $\cA$ is bifurcating. Then there are at least two possible responses in $\cY_\cH(x_T,S^h_{T-1}[\cA])$. Therefore $\cost(x_T,h'(x_T)) \geq \minsec(x_T)$. By the definition of $\sscr_\cost$, 
\[
\cost(x_T,h(x_T)) \leq \sscr_\cost \cdot \minsec(x_T).
\]
 Therefore 
\[
\cost(x_T,h(x_T)) \leq r_\cost \cost(x_T,h'(x_T))) = \gamma \cost(x_T,h'(x_T))).
\]
In both cases, 

\[
\cost(x_T,h(x_T)) - \cost(x_T,h'(x_T)) \leq (\gamma - 1)\cost(x_T,h'(x_T)).
\]

Therefore 

\[
\cost(S^h[\bar \cA]) \leq \cost(S_T^{h'}[\cA]) + (\gamma - 1)\cost(x_T,h'(x_T)) \leq \gamma\cost(S_T^{h'}[\cA]),
\]

where the last inequality follows since 
\[
\cost(S_T^{h'}[\cA])\leq \cost(S_T^{h'}[\cA]).
\]
Thus for all $h \in \bar{\cH}$, 
\[
\cost(S^h[\bar \cA]) \leq \gamma \cdot \cost(\cA), 
\]
hence $\psi \leq \gamma \cdot \cost(\cA)$. Combining this with \eqref{costbar}, the proof is concluded.
\end{proof}

In the proof of \thmref{worstcase} we further use the following lemmas.
\begin{lemma}\label{lem:greedylem}
Let $\beta,\alpha \geq 1$. Let $f,Q,\eta$ such that \asref{fterms} holds.
If for all $S \subseteq \pairs$, 
\begin{equation}\label{eq:greedycond}
\max_{x \in \cX} u^f(x,S) \geq \frac{Q- f(S)}{\beta\OPT},
\end{equation}
then for any $\alpha$-approximate greedy algorithm $\cA$ with $u^f$,
\[
\cost(\cA) \leq \alpha \beta (\ln(Q/\eta) + 1)\OPT.
\]
\end{lemma}

\begin{proof}
Let $h \in \cH$. Denote $S_t := S_t^h[\cA]$, and let $(x_t,y_t)$ be the action-response pair selected by the algorithm at iteration $t$, if $h^* = h$. 
Since $\cA$ is $\alpha$-approximately greedy with $u^f$, it follows from \eqref{greedycond} that $u(x_t,S_t) \geq (Q-f(S_t))/(\alpha\beta\OPT)$.

We have
\begin{align*}
\frac{f(S_t) - f(S_{t-1})}{\cost(x_t,y_t)} = \frac{\delta_f((x_t,y_t) \mid S_{t-1})}{\cost(x_t,y_t)} \geq u(x_t,S_{t-1}) \geq \frac{Q-f(S_{t-1})}{\alpha \beta \OPT}.
\end{align*}
hence

$Q - f(S_{t}) \leq (Q - f(S_{t-1}))(1- \frac{\cost(x_t,y_t)}{\alpha\beta\OPT}).$

Since $f(\emptyset) = 0$, it follows
\begin{align*}
 Q - f(S_{t}) &\leq Q\prod_{i \in [t]} (1-\frac{\cost(x_i,y_i)}{\alpha\beta\OPT})\\\
&\leq Q \exp(-\frac{1}{\alpha\beta\OPT}\sum_{i \in [t]}\cost(x_i,y_i)) \\
&= Q\exp(-\frac{\cost(S_t)}{\alpha\beta\OPT}).
\end{align*}
Let $T = |S^h[\cA]|$. Then $f(S_{T-1}) \leq Q-\eta$. Therefore
\[
\eta \leq Q - f(S_{T-1}) \leq Q \exp(-\cost(S_{T-1})/(\alpha\beta\OPT)).
\]
Hence $\cost(S_{T-1}^h[\cA]) \leq \alpha\beta \ln(Q/\eta)\OPT.$
Therefore
\begin{equation}\label{eq:costh}
\cost(S^h[\cA]) \leq \alpha\beta \ln(Q/\eta)\OPT + \cost(x_T,h(x_T)).
\end{equation}
By \eqref{greedycond},
\begin{equation}\label{eq:lastit}
u(x_T,S_{T-1}^h[\cA]) \geq (Q-f(S_{T-1}^h[\cA]))/(\alpha\beta\OPT).
\end{equation}
We have $f(S \cup \{(x_T,h(x_T)\}) \geq Q$, therefore 
\[
\delta_{\min(f,Q)}((x_T,h(x_T)) | S_{T-1}^h[\cA]) = Q-f(S_{T-1}^h[\cA]).
\]
 It follows that 
\[
u(x_T,S_{T-1}^h[\cA]) \leq \frac{Q-f(S_{T-1}^h[\cA])}{\cost(x_T,h(x_T))}.
\]
Combining this with \eqref{lastit}, we conclude that 
$\cost(x_T,h(x_T)) \leq \alpha\beta\OPT.$
Combining with \eqref{costh} and minimizing over $h \in \cH$, we conclude
that 
\[
\cost(\cA) \leq \alpha \beta (\ln(Q/\eta) + 1)\OPT.
\]

\end{proof}

\begin{lemma}\label{lem:fprime}
Let $f,Q,\eta$ such that \asref{fterms} holds and $f$ is consistency-aware. 
Let $S \subseteq \pairs$. Define $f':2^\pairs \rightarrow \reals_+$ by
$f'(T) := f(T\cup S) - f(S)$. Let $Q' = Q -  f(S)$.
Then 
\begin{enumerate}
\item $f'$ is submodular, monotone and consistency-aware, with $f'(\emptyset) = 0$. 
\item Let $\cA$ be an interactive algorithm for $f',Q'$. Let $\beta \geq 1$.
If 
\begin{equation}\label{eq:optempty}
\max_{x\in \cX} u^{f'}(x,\emptyset) \geq \frac{Q'}{\beta \OPT'},
\end{equation}
where $\OPT'$ is the optimal cost for $f',Q'$, then 
\[
\max_{x\in \cX} u^f(x,S) \geq \frac{Q-f(S)}{\beta\OPT}.
\]
\end{enumerate}
\end{lemma}
\begin{proof}
First, we prove the claim for the case $f(S) \leq Q$. 
By the monotonicity of $f$, any interactive algorithm that obtains $f(T) \geq Q$ obtains also $f'(T) \geq Q'$. Denote the optimal cost for $f',Q'$ by $\OPT'$. Then $\OPT' \leq \OPT$. Since $f$ is submodular, so is $f'$.
Further, since $f$ is consistency-aware with $Q$, for $T$ which is inconsistent with $\cH$ we have
\[
f'(T) = f(T\cup S) - f(S) \geq Q - f(S) = Q'.
\]
Hence $f'$ is consistency-aware with $Q'$. 
Now suppose that \eqref{optempty} holds, then
\begin{equation}\label{eq:optempty2}
u^{f'}(x,\emptyset) \geq \frac{Q'}{\beta \OPT'} = \frac{(Q-f(S))}{\beta \OPT'} \geq \frac{(Q-f(S))}{\beta\OPT}.
\end{equation}

We have
\[
u^{f'}(x,\emptyset) := \min_{h \in \vs(S)}\frac{\delta_{\min(f',Q')}((x,h(x)) \mid \emptyset)}{\cost(x,h(x))}.
\]
For any $(x,y)$,  $\delta_{\min(f',Q')}((x,y) \mid \emptyset) = \min\{f'(\{(x,y)\}),Q'\}.$ 
Since $f \leq Q$, we have $f'(\{(x,y)\}) = f(\{(x,y)\}\cup S) - f(S) \leq Q - f(S) = Q'.$ Hence 
\begin{align*}
\delta_{\min(f',Q')}((x,y) \mid \emptyset) = f(\{(x,y)\}\cup S) - f(S) = \delta_f((x,y) \mid S) = \delta_{\min(f,Q)}((x,y) \mid S).
\end{align*}
Therefore
$u^{f'}(x,\emptyset) = u^f(x,S)$. 
We conclude from \eqref{optempty2} that if $f(S) \leq Q$, then 
\[
u^{f}(x,S) \geq \frac{Q-f(S)}{\beta \OPT}.
\]

To finalize the proof, if $f(S) \leq Q$ does not hold, 
 consider $\bar f := \min(f,Q)$. Since $f$ is submodular, so is $\bar f$
  \citep{Narayanan97}. All other properties assumed for $f$ are also preserved
  by $\bar f$, and $u^f \equiv u^{\bar f}$. Therefore
\[
u^{f}(x,S) = u^{\bar f}(x,S) \geq \frac{Q-\bar f(S)}{\beta \OPT} \geq \frac{Q-f(S)}{\beta \OPT}.
\]
\end{proof}

Using the lemmas above, \thmref{worstcase} is easily proved.
\begin{proof}[Proof of \thmref{worstcase}]
Fix $S \subseteq \pairs$, and let $f',Q',\OPT'$ be as in \lemref{fprime}. Let $\cA^*$ be an optimal algorithm for $f',Q'$. Since $f$ is a learning objective, then so is $f'$, and by \lemref{bifurcating} we can choose $\cA^*$ to be bifurcating. Combining this with the first part of \lemref{fprime}, the conditions of \lemref{onestep} hold for $f',Q'$. Therefore
\[
\max_{x \in \cX} u^{f'}(x,\emptyset) \geq Q'/\cost(\cA^*) \geq Q'/(\sscr_\cost\cdot\OPT').
\]
By the second part of \lemref{fprime}, 
\[
\max_{x \in \cX} u^{f}(x,S) \geq \frac{Q-f(S)}{\sscr_\cost\cdot\OPT}.
\]
This holds for any $S \subseteq \pairs$. 
Therefore, by \lemref{greedylem}, 
\[
\cost(\cA) \leq \alpha (\ln(Q/\eta) + 1)\cdot \sscr_\cost \cdot \OPT.
\]
\end{proof}
The approximation bound depends linearly on $\sscr_\cost$. 
In the next section, we show that such a linear dependence is necessary for any greedy algorithm for learning objectives.

\subsection{A lower bound for learning objectives}\label{sec:learninglower}

In this section we study the limitations of greedy algorithms for the interactive selection problem with response-dependent costs. Thus, we are interested in lower bounds that hold for all greedy algorithms, regardless of their utility function. However, for any fixed $\cX$ there exists a tailored utility function $u_\cX$ that induces the optimal action-selection behavior: this is the utility function which gives the maximal value to the next action that should be selected, based on the optimal selection path for $\cX$. 

Since we are interested in general strategies for greedy selection, and not in ones that are tailored for a single specific action set $\cX$, we study the performance of a greedy algorithm on a \emph{family} of problems, each with a possibly different action set $\cX$. The other problem parameters $f,\cH,\cost,\cY$ are the same in all the problems in the family. The approximation factor of a greedy algorithm for a given family is its worst-case factor over all the problems in the family.

Formally, define \emph{local} greedy algorithms as follows. Assume there is a super-domain of all possible actions $\bar{\cX}$, and consider an algorithm which receives as input a subset $\cX \subseteq \bar{\cX}$ of available actions. We say that such an algorithm is \emph{local} greedy if it greedily selects the next action out of $\cX$ using a fixed utility function $u: \bar{\cX} \times 2^{\bar{\cX}\times \cY} \rightarrow \reals_+$, which does not depend on $\cX$.
The following lower bound shows that there exists a learning objective such that the approximation guarantee of any local greedy algorithm grows with $\sscr_\cost$ or is trivially bad. 
\begin{theorem}\label{thm:lowerboundlearning}
Let $f$ be the version-space reduction objective function with the corresponding $Q = 1-1/|\cH|$ and $\eta = 1/|\cH|$. 
For any value of $\OPT$, $\sscr_\cost > 1$, and any integer value of $Q/\eta$, there exist $\bar{\cX},\cH$, and $\cost$ such that $\cost(x,y)$ depends only on $y$, and such that for any local greedy algorithm $\cA$, there exists an input domain $\cX\subseteq \bar{\cX}$ such that, for $\eta$ as in \thmref{worstcase},
\[
\cost(\cA) \geq \min\left(\frac{\sscr_\cost}{\log_2(Q/\eta)},\frac{Q/\eta}{\log_2(Q/\eta)}\right) \cdot \OPT.
\]
Here $\cost(\cA)$ and $\OPT$ refer to the costs for the domain $\cX$.
\end{theorem}

\begin{proof}
Define $\cY = \{1,2,3\}$. Set $k = Q/\eta$, and define $\cH = \{ h_i \mid i \in [k]\}$, where $h_i$ will be given below. Note that for the version-space reduction objective, $Q = |\cH|$, $\eta = 1$, so indeed $Q/\eta = k$. 
Let 
\[
\bar{\cX} = \{a_i \mid i \in [k]\} \cup \{b_j^t \mid j\in [k], t \in [\ceil{\log_2(k-2)}]\}.
\]
Set $c_1,c_2,c_3$ such that $c_1 = 0$, $c_2 > 0$, and $c_3 = c_2\sscr_\cost$. Let $\cost(x,y) = c_y$ for all $x \in \cX$. 
Define each $h_i$ as follows: for $a_j$,
\[
h_i(a_j) := \begin{cases}
1 & i = j \\
2 & i \neq j.
\end{cases} 
\]
For $b^t_j$ and $i \neq j$, let $l_{i,j}$ be the location of $i$ in $(1,\ldots,j-1,j+1,\ldots,k)$, where the locations range from $0$ to $k-2$. Denote by $l^t_{i,j}$ the $t$'th most significant bit in the binary expansion of $l_{i,j}$ to $\ceil{\log_2(k-2)}$ bits. Define
\[
h_i(b^t_j) := \begin{cases}
1 & i \neq j \wedge l_{i,j}^t = 0\\
2 & i \neq j \wedge l_{i,j}^t = 1\\
3 & i = j
\end{cases} 
\]
See \tabref{costslearn} for illustration.

\tablecostslearn

Fix an index $n \in [k]$. Let 
\[
\cX_n = \{a_i \mid i \in [k]\} \cup \{ b^t_n \mid t \in [\ceil{\log_2(k-2)}]\}.
\]

To upper bound $\OPT$, we now show an interactive algorithm for $\cX_n$ and bound its worst-case cost. 
On the first iteration, the algorithm selects action $a_n$. If the result is $1$, then $\vs(S) = \{h_n\}$, hence $f(S) \geq Q$. In this case the cost is $c_1 = 0$. Otherwise, the algorithm selects all actions in $\{b^t_n \mid  t \in [\ceil{\log_2(k-2)}\}$. The responses reveal the binary expansion of $l_{j,n}$, thus limiting the version space to a single $h_i$, hence $f(S) \geq Q$. In this case the total cost is at most $c_2 \ceil{\log_2(k-2)}$.

Now, consider a local greedy algorithm with some utility function $u$. 
Let $\sigma:[k] \rightarrow [k]$ be a permutation that represents the order in which $a_1,\ldots,a_k$ would be selected by the utility function if only $a_i$ were available, and their response was always $2$. Formally,\footnote{We may assume without loss of generality that $u(x,S) = 0$ whenever $(x,y) \in S$.} 
\[
\sigma(i) = \argmax_{i \in [k]} u(a_{\sigma(i)},\{(a_{\sigma(i')},2) \mid i' \in [i-1] \}).
\]

Suppose the input to the algorithm is $\cX_{\sigma(k)}$.
Denote 
\[
S_i = \{(a_{\sigma(i')},2) \mid i' \in [i-1] \}, 
\]
and suppose $h^* = h_{\sigma(k)}$. First, assume that 
\begin{equation}\label{eq:assume}
\max_t u(b^t_{\sigma(k)},S_{i'-1}) < u(a_{\sigma(k)},S_{k-1}).
\end{equation}
Then all of $a_{\sigma(1)},\ldots,a_{\sigma(k-1)}$ are selected before any of $b^t_{\sigma(k)}$, and the version space is reduced to a singleton only after these $k-1$ actions. Therefore the cost of the run is at least $c_2(k-1)$. 
Second, assume that \eqref{assume} does not hold. Then there exists an integer $i'$ such that 
\[
\max_t u(b^t_{\sigma(k)},S_{i'-1}) > u(a_{\sigma(i)},S_{i'-1}).
\]
 Let $i'$ be the smallest such integer. Then, the algorithm receives $2$ on each of the actions $a_{\sigma(1)},\ldots,a_{\sigma(i'-1)}$, and its next action is $b^t_{\sigma(k)}$ for some $t$. Hence the cost of the run is at least $c_3.$

To summarize, the worst-case cost of every local greedy algorithm is at least $\min\{c_3, c_2(k-1)\}$ for at least one of the inputs $\cX_n$, while $\OPT$ is for any $\cX_n$ at most $c_2 \ceil{\log_2(k-2)}$. The statement of the theorem follows.
\end{proof}

The lower bound above matches the upper bound in \thmref{worstcase} in terms of the linear dependence on $\sscr_\cost$, but not in terms of the dependence on $Q/\eta$, which is $O(\ln(Q/\eta))$ in the upper bound. Nonetheless, a lower bound of $\Omega(\ln(Q/\eta))$ is known for any efficient algorithm, even for the simpler setting without response-dependent costs and without interaction \citep{Wolsey82,Feige98}. In particular, this lower bound holds for any greedy algorithm with an efficiently computable utility function.

\subsection{An approximation upper bound for general objectives}\label{sec:generalobj}

We now turn to consider general objectives. We showed above that for learning objectives, the achievable approximation guarantee for greedy algorithms is characterized by $\sscr_\cost$. We now turn to general consistency-aware objective functions. We show that the factor of approximation for this class depends on a different property of the cost function, which is lower bounded by $\sscr_\cost$. Define 
\[
\cmax := \max_{(x,y) \in \cX\times \cY} \cost(x,y).
\]
Recall that $\minsec(x)$ is the second-smallest cost for $x$, and let
\[
\phi_{\min} := \min_{x \in \cX} \minsec(x), \qquad \gsscr_\cost := \frac{\cmax}{\phi_{\min}}.
\]
We term the ratio $\gsscr_\cost$ the \emph{Global second smallest cost ratio}.
As we show below, the approximation factor is best when $\gsscr_\cost$ is equal to $1$. This is the case if there is at most one preferred response for every action, and in addition, all the non-preferred responses for all actions have the same cost. 

\begin{theorem}\label{thm:nonlearning}
Let $f:2^\pairs \rightarrow \reals_+,Q >0,\eta > 0$ such that \asref{fterms} holds and $f$ is consistency-aware. 
Let $\cA$ be an $\alpha$-approximate greedy algorithm for the utility function $u^f$. Then
\[
\cost(\cA) \leq 2\min(\gsscr_\cost,\crr_\cost) \cdot \alpha \cdot (\ln(Q/\eta)+1) \cdot \OPT. 
\]
\end{theorem}
Similarly to \thmref{worstcase} for learning objectives, this result for general objectives is a significant improvement over the trivial bound, mentioned in \secref{setting}, which depends on the cost ratio, since the ratio $\gsscr_\cost/\crr_\cost$ can be unbounded. For instance, consider a case where each action has one response with a cost of $1$ and all other responses have a cost of $M \gg 1$. Then $\crr_\cost = M$ but $\gsscr_\cost = 1$.

The proof of \thmref{nonlearning} hinges on two main observations: First, any interactive algorithm may be ``reordered'' without increasing its cost, so that all actions with only one possible response (given the history so far) are last. Second, there are two distinct cases for the optimal algorithm: In one case, for all $h \in \cH$, the optimal algorithm obtains a value of at least $Q/2$ before performing actions with a single possible response. In the other case, there exists at least one mapping $h$ for which actions with a single possible response obtain at least $Q/2$ of the value. We start with the following lemma, which handles the case where $\OPT < \phi_{\min}$. 

\newcommand{\withone}[1]{{#1}^{[y(x)]}}
\begin{lemma}\label{lem:smallopt}
Let $f:2^\pairs \rightarrow \reals_+$, $Q > 0$. Suppose that $f$ is submodular, and $f(0) = \emptyset$.
If $\OPT < \phi_{\min}$, then 
\[
\max_{x \in \cX}u^f(x,\emptyset) \geq Q/OPT.
\]
\end{lemma}
\begin{proof}
For every action $x \in \cX$ there is at most a single $y$ with $\cost(x,y) < \phi_{\min}$. Denote this response by $y(x)$. 
Let $\cA$ be an optimal algorithm for $f,Q$. For any value of $h^* \in \cH$, $\cA$ only receives responses with costs less than $\phi_{\min}$. Therefore for any $x$ that $\cA$ selects, it receives the response $y(x)$, regardless of the identity of $h^*$. In other words, for all $h \in \cH$, in every iteration $t$, $\cA$ selects an action $x$ such that 
\[
\cY(x,S^h_{t-1}[\cA]) = \{y(x)\}.
\]
 It follows that for all $t$, $S^h_{t}[\cA]$ is the same for all $h \in \cH$.
Therefore, there is a fixed set of actions that $\cA$ selects during its run, regardless of $h^*$. Let $\cX' \subseteq \cX$ be that set. Then for all $h \in \cH,x\in X'$, $h(x) = y(x)$. 
For a set $A \subseteq \cX$, denote 
\[
\withone{A} = \{(x,y(x)) \mid x \in A\}.
\]
 
We have $f(\withone{\cX'}) \geq Q$ and $\cost(\withone{\cX'}) = \OPT$.
By the submodularity of $f$, and since $f(\emptyset) = 0$, we have

\[
Q/\OPT \leq f(\withone{\cX'})/\OPT \leq \sum_{x \in \cX'} f((x,y(x)))/\sum_{x \in \cX'}\cost(x,y(x)).
\]

Therefore there exists some $x \in \cX'$ with 
\[
f((x,y(x)))/\cost(x,y(x)) \geq Q/\OPT.
\]
Moreover, for this $x$ we have $\cY(x,\emptyset) = \{y(x)\}$. Therefore 
\[
u^f(x,\emptyset) = f((x,y(x)))/\cost(x,y(x)) \geq Q/\OPT.
\]
\end{proof}

We now turn to the main lemma, to address the two cases described above.
\begin{lemma}\label{lem:optc2}
Let $f:2^\pairs \rightarrow \reals_+$, $Q > 0$. Suppose that $f$ is submodular, and $f(0) = \emptyset$. Assume that $f$ is consistency-aware.
There exists $x \in \cX$ such that 
\[
u^f(x,\emptyset) \geq \frac{Q}{2\min(\gsscr_\cost, \crr_\cost) \OPT}.
\]

\end{lemma}

\begin{proof}
If $\OPT < \phi_{\min}$, the statement holds by \lemref{smallopt}. 
Suppose that $\OPT \geq \phi_{\min}$. Let $\cA^*$ be an optimal algorithm for $f,Q$. We may assume without loss of generality, that for any $h^* \in \cH$, if $\cA^*$ selects an action that has only one possible response (given the current version space) at some iteration $t$, then all actions selected after iteration $t$ also have only one possible response.
This does not lose generality: let $t$ be the first iteration such that the action at iteration $t$ has one possible response, and the action at iteration $t+1$ has two possible responses. Consider an algorithm which behaves the same as $\cA^*$, except that at iteration $t$ it selects the second action, and at iteration $t+1$ it selects the first action (regardless of the response to the first action).  This algorithm has the same cost as $\cA^*$.

For $h \in \cH$, define $\val(h) := f(S_{t_h}^h[\cA^*])$, where $t_h$ is the last iteration in which an action with more than one possible response (given the current version space) is selected, if $h^* = h$. 
Consider two cases: 
\begin{enumerate}[(a)]
\item $\min_{h \in \cH} \val(h) \geq Q/2$ and 
\item $\exists h \in \cH, \val(h) < Q/2$. 
\end{enumerate}
In case (a), there is a bifurcating algorithm that obtains  $f(S) \geq Q/2$ at cost at most $\OPT$: This is the algorithm that selects the same actions as $\cA^*$, but terminates before selecting the first action that has a single response given the current version space. We also have $\sscr_\cost \leq \min(\gsscr_\cost,\crr_\cost)$. By \lemref{onestep}, there exists some $x \in \cX$ such that 
\[
u^f(x,\emptyset) \geq \frac{Q}{2\min(\gsscr_\cost,\crr_\cost)\OPT}.
\]

In case (b), let $h \in \cH$ such that $\val(h) < Q/2$. Denote $S_t := S_t^h[\cA^*]$. Let $(x_t,h(x_t))$ be the action and the response received in iteration $t$ if $h^* = h$. Then $f(S_{t_h}) < Q/2$. 
Let $S' = \{(x_t,h(x_t)) \mid t > t_h\}$. Then $f(S_{t_h}\cup S') \geq Q$. Since $f(\emptyset) = 0$ and $f$ is submodular,
\[
f(S') = f(S') - f(\emptyset) \geq f(S_{t_h}\cup S') - f(S_{t_h}) \geq Q - \val(h) \geq Q/2.
\]
In addition, $f(S') \leq \sum_{t> t_h} f(\{(x_t,h(x_t))\})$. Hence 
\[
\frac{Q}{2\OPT} \leq \frac{f(S')}{\OPT} \leq \frac{\sum_{t> t_h} f(\{(x_t,h(x_t))\})}{\sum_{t> t_h}\cost(x_t,y_t)}.
\]
Therefore there is some $t'$ such that 

\[
\frac{f(\{(x_{t'},h(x_{t'}))\})}{\cost(x_{t'},h(x_{t'}))} \geq \frac{Q}{2\OPT}.
\]

Therefore,
\begin{align*}
u^f(x_{t'},\emptyset) &= \min_{y \in \cY(x_{t'},\emptyset)} \min\{f(\{(x_{t'},y)\}),Q\}/\cost(x_{t'},y) \\
&\geq \min\{Q/\cmax,\min_{y \in \cY}f(\{(x_{t'},y)\})/\cost(x_{t'},y)\}\\
  &\geq \min\{ Q/\cmax, \frac{Q}{2\OPT}, \min_{y \in \cY\setminus \{h(x_{t'})\}} f(\{(x_{t'},y)\})/\cost(x_{t'},y)\}.
\end{align*}

Now, 
\[
\cmax = \gsscr_\cost \cdot \phi_{\min} \leq \gsscr_\cost\cdot \OPT, 
\]

from our assumption that $\OPT \geq \phi_{\min}$. Also
\[
\cmax \leq \crr_\cost \cost(x_{t'},h(x_{t'})) \leq \crr_\cost \cdot \OPT.
\]

Therefore
\begin{align*}
u^f(x_{t'},\emptyset) \geq \min\Bigg\{&\frac{Q}{2\min(\gsscr_\cost,\crr_\cost)\OPT}, \min_{y \in \cY\setminus \{h(x_{t'})\}} \frac{f(\{(x_{t'},y)\})}{\cost(x_{t'},y)}\Bigg\}.
\end{align*}

We have left to show a lower bound on 
\[
\min_{y \in \cY\setminus \{h(x_{t'})\}} \frac{f(\{(x_{t'},y)\})}{\cost(x_{t'},y)}.
\]
 By the choice of $t'$, $x_{t'}$ has only one possible response given the current version space, that is $|\cY(x_{t'},S_{{t'}-1})| = 1$. Since the same holds for all $t > t_h$, we have $\vs(S_{t'-1}) = \vs(S_{t_h})$, hence also $\cY(x_{t'},S_{t_h}) = \{h(x_{t'})\}$. It follows that for $y \in \cY\setminus \{h(x_{t'})\}$, the set $S_{t_h} \cup \{(x_{t'},y)\}$  is not consistent with any $h \in \cH$.
Since $f$ is consistency-aware, it follows that $f(S_{t_h} \cup \{(x_{t'},y)\}) \geq Q$. Therefore 
\[
f(\{(x_{t'}, y)\}) = f(\{(x_{t'}, y)\}) - f(\emptyset) \geq f(S_{t_h} \cup \{(x_{t'},y)\}) - f(S_{t_h}) \geq Q - \val(h) \geq Q/2. 
\]
Hence 
\[
\frac{f(\{(x_{t'}, y)\})}{\cost(x_{t'}, y)} \geq \frac{Q}{2\cmax} \geq \frac{Q}{2\min(\gsscr_\cost,\crr_\cost)\OPT}.
\]
It follows that 
\[
u^f(x_t,\emptyset) \geq \frac{Q}{2\cmax} \geq \frac{Q}{2\min(\gsscr_\cost,\crr_\cost)\OPT} 
\]
also in case (b).
\end{proof}

Using the lemmas above, the proof of \thmref{nonlearning} is straightforward.

\begin{proof}[Proof of \thmref{nonlearning}]

Fix $S \subseteq \pairs$, and let $f',Q',\OPT'$ as in \lemref{fprime}. Let $\cA^*$ be an optimal algorithm for $f',Q'$. 
From the first part of \lemref{fprime}, the conditions of \lemref{optc2} hold for $f',Q'$. Therefore
\[
\max_{x \in \cX} u^{f'}(x,\emptyset) \geq \frac{Q'}{2\min(\gsscr_\cost,\crr_\cost)\OPT'}.
\]
By the second part of \lemref{fprime},
\[
u^{f}(x,S)  \geq \frac{Q-f(S)}{2\min(\gsscr_\cost,\crr_\cost)\OPT}.
\]
This holds for any $S \subseteq \pairs$. 
Therefore, by \lemref{greedylem},
\[
\cost(\cA) \leq 2 \alpha \min(\gsscr_\cost,\crr_\cost) (\ln(Q/\eta) + 1)\cdot \OPT.
\]
\end{proof}

The guarantee of \thmref{nonlearning} for general objectives is weaker than the guarantee for learning objectives given in \thmref{worstcase}: The ratio between the terms, $\min(\gsscr_\cost,\crr_\cost)/\sscr_\cost$, is always at least $1$, and can be unbounded. For instance, if there are two actions that have two responses each, and all action-response pairs cost $1$, except for one action-response pair which costs $M \gg 1$, then $\sscr_\cost = 1$ but $\crr_\cost = \gsscr_\cost = M$. Nonetheless, in the following section we show that for general functions, a dependence on $\min(\gsscr_\cost,\crr_\cost)$ is unavoidable in any greedy algorithm. 

\subsection{A lower bound for general functions}\label{sec:generallower}

The following lower bound holds for any local greedy algorithm for general functions.
\begin{theorem}\label{thm:lowerboundgeneral}
For any values of $\gsscr_\cost,\crr_\cost > 0$, there exist $\bar{\cX},\cY,\cH,\cost$ with $|\cY| = 2$ and $\sscr_\cost = 1$, and a submodular monotone $f$ which is consistency-aware, with $Q/\eta = 1$, such that for any local greedy algorithm $\cA$, there exists an input domain $\cX\subseteq \bar{\cX}$ such that
\[
\cost(\cA) \geq \half \min(\gsscr_\cost,\crr_\cost)\cdot \OPT,
\]
where $\cost(\cA)$ and $\OPT$ refer to the costs of an algorithm running on the domain $\cX$.
\end{theorem}

\begin{proof}
Define $\cY := \{0,1\}$. Let $g,r >0$ be the desired values for $\gsscr_\cost, \crr_\cost$. Let $c_1 > 0$, $c_2 := c_1\min(g,r)$. If $g < r$, define $c_3 := c_1/r$, $c_4 := c_1$. Otherwise, set $c_4 := c_3 := c_2/g$.
Define $k := \ceil{c_2/c_1} + 1$. 
Let 
\[
\bar{\cX} = \{a_i \mid i \in [k]\} \cup \{b_i \mid i \in [k]\} \cup \{c\}.
\]
 Let $\bar{\cH} := \{h_i \mid i \in [k]\}$, where $h_i$ is defined as follows:
\begin{align*}
&\forall i,j \in [k], h_i(a_j) = h_i(b_j) = \one[i = j], \\
&\forall i \in [k], h_i(c) = i \bmod 2.
\end{align*}
Let the cost function be as follows, where $c_2 \geq c_1 > 0$, and $c_3,c_4 > 0$: $\cost(a_i,y) = c_1$, $\cost(b_i,y) = c_{y+1}$, and $\cost(c, y) = c_{y+3}.$
Then $\gsscr_\cost = g$, $\crr_\cost = r$ as desired. See \tabref{costsgen} for an illustration.

\tablecostsgen

Define $f$ such that $\forall S \subseteq \pairs$, $f(S) = Q$ if there exists in $S$ at least one of $(a_i,1)$ for some $i \in [k]$ or $(b_i,y)$ for some $i \in [k],y\in \cY$. Otherwise, $f(S) = 0$. Note that $(f,Q)$ is consistency-aware.

Fix an index $n \in [k]$. Let $\cX_n = \{a_i \mid i \in [k]\} \cup \{b_n\}$.
We have $\OPT = 2c_1$: An interactive algorithm can first select $a_n$, and then, only if the response is $y=0$, select $b_n$. 
Now, consider a local greedy algorithm with a utility function $u$.
 Let $\sigma:[k] \rightarrow [k]$ be a permutation that represents the order in which $a_1,\ldots,a_k$ would be selected by the utility function if only $a_i$ were considered, and their response was always $y=0$. Formally, \footnote{We may assume without loss of generality that $u(x,S) = 0$ whenever $(x,y) \in S$.}
\[
\sigma(i) = \argmax_{i \in [k]} u(a_{\sigma(i)},\{(a_{\sigma(i')},0) \mid i' \in [i-1] \}).
\]

Now, suppose the input to the algorithm is $\cX_{\sigma(k)}$.
Denote 
\[
S_i = \{(a_{\sigma(i')},0) \mid i' \in [i-1] \}.
\]
Suppose that there exists an integer $i'$ such that $u(b_{\sigma(k)},S_{i'-1}) > u(a_{\sigma(i)},S_{i'-1})$, and let $i'$ be the smallest such integer. Then, if the algorithm receives $0$ on each of the actions $a_{\sigma(1)},\ldots,a_{\sigma(i'-1)}$, its next action will be $b_{\sigma(k)}$. 
In this case, if $h^* = h_{\sigma(k)}$, then $b_{\sigma(k)}$ is queried before $a_{\sigma(k)}$ is queried and the response $y=1$ is received. Thus the algorithm pays at least $c_2$ in the worst-case. 

On the other hand, if such an integer $i'$ does not exist, then if $h^* = h_{\sigma(k)}$, the algorithm selects actions $a_{\sigma(1)},\ldots,a_{\sigma(k-1)}$ before terminating. In this case the algorithm receives $k-1$ responses $0$, thus its cost is at least $c_1(k-1)$. 
To summarize, every local greedy algorithm pays at least $\min\{c_2, c_1(k-1)\}$ for at least one of the inputs $\cX_n$, while $\OPT = 2c_1$. By the definition of $k$,  $\min\{c_2, c_1(k-1)\} \geq c_2$. 
Hence the cost of the local greedy algorithm is at least $\frac{c_2}{2c_1}\OPT$. 
\end{proof}

To summarize, for both learning objectives and general objectives, we have shown that the factors $\sscr_\cost$ and $\gsscr_\cost$, respectively, characterize the approximation factors obtainable by a greedy algorithm.

\section{Experiments}\label{sec:exp}

\begin{table}[th]
\tableresults
\caption{Results of experiments. }
\label{tab:results}
\end{table}

We performed experiments to compare the worst-case costs of a greedy algorithm that uses the proposed $u^f$, to a greedy algorithm that ignores response-dependent costs, and uses instead a variant of $u^f$, notated $u_2^f$, that assumes that responses for the same action have the same cost, which was set to be the maximal response cost for this action. We also compared to $u_3^f$, a utility function which gives the same approximation guarantees as given in \thmref{nonlearning} for $u^f$.
Formally, 
\[
u_2^f(x,S) := \min_{h \in \vs(S)}\frac{\delta_{\min(f,Q)}((x,h(x)) \mid S)}{\max_{y \in \cY}\cost(x,y)}
\]
and 
\[
u_3^f(x,S) := \min_{h \in \vs(S)}\frac{\delta_{\min(f,Q)}((x,h(x)) \mid S)}{\min\{\cost(x,h(x)), \phi_{\min}\}}.
\]

It can be easily shown that for general objectives with $\gsscr_\cost \leq \crr_\cost$, the utility function $u_3^f$ has the same approximation guarantees as given in \thmref{nonlearning} for $u^f$, by observing that $u^f_3$ is equal to $u^f$ for $\cost'(x,y) := \min\{\cost(x,y), \phi_{\min}\}$ and that the optimal value for $\cost$ is at most $\gsscr_\cost$ times the optimal value for $\cost'$. 
Thus it is instructive to compare these approaches in practice.

We tested these algorithms on a social network marketing objective, where users in a social network are partitioned into communities.
Actions are users, and a response identifies the community the user belongs to. 
We tested two objective functions. The first objective is ``edge users'', which counts how many of the actions are users who have at least one friend not from their community, assuming that these users can be valuable promoters across communities. The definition of this objective function is given in \secref{example}. The target value $Q$ was set to $50$, that is, the goal was to find $50$ users with friends in a different community.
The second objective function was the version-space reduction function, and the goal was to identify the true partition into communities out of the set of possible partitions.

In each experiment, a hypothesis class $\cH$, representing the set of possible partitions into
communities, was generated as follows: Given a set of users $A \subseteq \cX$ of size $k$,
define the hypothesis $h_A$, which induces a partition of the users in the
graph into communities, by setting the users in $A = (x_1,\ldots,x_k)$
to be ``center users'', and defining the community centered
around user $x_i$ as all the users in the social network that are closer to
user $x_i$ than to any of the other users in $A$. Here, the distance between
two users is the number of edges in the shortest path between these users in the social
network graph. Formally, $h_A(x) = \argmin_{i \leq k} d(x,x_i)$, where $d(x,x_i)$
is the shortest-path distance between $x$ and $x_i$, and ties are broken arbitrarily.
For each experiment reported below, $\cH$ was set by selecting $k'$ sets $A_1,\ldots,A_{k'}$, each set of size $k$, uniformly at random from the users in the network, and setting $\cH := \{h_{A_1},\ldots,h_{A_{k'}}\}$. In our experiments we generated hypothesis classes $\cH$ according to the combinations $k = 3, k'=100$ and $k = 10, k' = 500$.

We report the worst-case cost $\cost(\cA)$ for each of the problems we tested. 
We compared the worst-case costs of the algorithms under several configurations of number of communities and the values of $\sscr_\cost$, $\gsscr_\cost$. The cost ratio $\crr_\cost$ was infinity in all experiments, obtained by always setting a single response to have a cost of zero for each action. Social network graphs were taken from a friend graph from Facebook\footnote{\url{http://snap.stanford.edu/data/egonets-Facebook.html}} \citep{LeskovecMc12}, and a collaboration graph from Arxiv GR-QC community\footnote{\url{http://snap.stanford.edu/data/ca-GrQc.html}}~\mbox{\citep{LeskovecKlFa07}}. The results are reported in \tabref{results}. 

The results show an overall preference to the proposed $u^f$. 
It should not be surprising that $u_2^f$ performs poorly compared to $u^f$: this utility function always assumes the worst cost for each action, without taking into account the ratio between the improvement and the cost, or the current version space. Thus, a greedy algorithm that uses it is overly pessimistic, and might avoid certain actions even though in the current version space they cannot be very expensive. On the other hand, $u_3^f$ is too optimistic: it only considers the smallest and second-smallest possible costs when selecting an action. Thus, it does not differentiate between an action with some high costs, and an action with no high costs, even though in this case the latter is never worse than the former. Thus, both utility functions ignore cost information which $u^f$ takes into account, aiding it to obtain superior performance.

\section{Conclusions}\label{sec:conclusion}

In this work we analyzed the properties of a natural greedy algorithm for response-dependent costs, and showed that its approximation factor is significantly better than those trivially derived from previous results. We further showed that these guarantees cannot be significantly improved using a greedy algorithm, both for learning objectives and for general objectives. 

An important open problem is whether there exists an efficient, non-greedy algorithm, that can obtain even better approximation guarantees, especially in cases where $\sscr_\cost$ or $\gsscr_\cost$ are very large. Another question is whether similar guarantees can be obtained for the setting of average-case costs. We aim to study these questions in future work.

\section*{Acknowledgements}
This work was supported in part by the Israel Science Foundation (grant No. 555/15).

\section*{References}

\bibliography{shared.bib}

\end{document}